\documentclass{article}

\usepackage[nonatbib, final]{nips_2017}
\usepackage{subfig}
\usepackage{enumitem}  

\usepackage[utf8]{inputenc} 
\usepackage[T1]{fontenc}    
\usepackage{url}            
\usepackage{booktabs}       
\usepackage{amsfonts}       
\usepackage{nicefrac}       
\usepackage{microtype}      
\usepackage{amsmath}		
\usepackage{graphicx}
\usepackage{caption}
\usepackage{url}  
\usepackage{float}  

\begin{document}
\title{Detection and segmentation of the Left Ventricle in Cardiac MRI using Deep Learning}

%

\author{
  Alexandre Attia\\
  MVA \\
  ENS Paris-Saclay\\
  \texttt{alexandre.attia@ens-paris-saclay.fr} \\\And
  Sharone Dayan   \\
  MVA \\
  ENS Paris-Saclay\\
  \texttt{sharone.dayan@ens-paris-saclay.fr} \\
}

\maketitle

\begin{abstract} 
Manual segmentation of the Left Ventricle (LV) is a tedious and meticulous task that can vary depending on the patient, the Magnetic Resonance Images (MRI) cuts and the experts. Still today, we consider manual delineation done by experts as being the ground truth for cardiac diagnosticians. Thus, we are reviewing the paper - written by Avendi and al. - who presents a combined approach with Convolutional Neural Networks, Stacked Auto-Encoders and Deformable Models, to try and automate the segmentation while performing more accurately. Furthermore, we have implemented parts of the paper (around three quarts) and experimented both the original method and slightly modified versions when changing the architecture and the parameters. 
\end{abstract}

\section{Introduction}

The main purpose of this project is to review and deeply understand a paper from Avendi et al. [1] on automatic segmentation of the Left Ventricle using MRI cuts and their corresponding manual delineation done by experts, then we will try to challenge it. Segmentation of the left ventricle (LV) is currently done manually by experts (mainly junior surgeons doing delineation) from cardiac MRI images. When considering the intra-inter variance between the experts contours annotations and the heterogeneity in the MR images (illumination, volume effect, form differences depending on the pathology, etc.), the advantages of such an approach appear clearly. \\
The segmentation consists in assigning a label (ie a class e.g. LV or not LV) for each pixel. In oder to do so, we are relying on the recent advances in Deep Learning and Neural Networks, since Convolutional Neural networks (CNN) understand complex and interesting shape (such as organs or tissues), and hence are efficient at isolating a region of interest in the original image. Stacked auto-encoders operate afterwards as they provide different representations as to detect and segment the shape within the region of interest, which is in our case the LV. The uniqueness of this paper stands in its combined approach with Deformable models. Once we have inferred the shape of the LV in the priorly determined region of interest, Deformable models become very handy to prevent shrinkage, leakage and for 3D alignment to get the final accurate segmentation.\\

In the next parts, we will explain the methodology of Avendi et al., and present the parts of the paper that we have implemented. We will also try to challenge the paper by modifying slightly the architecture and tuning some parameters. For the implementation, we used Keras (minimalist Python library for deep learning) on top of Tensorflow backend. Code available at : \url{github.com/alexattia/Medical-Image-Analysis}.

\newpage
\section{Materials \& methods}
\subsection{Dataset \& preprocessing}
In order to understand the paper method and to implement parts of it, we used the same database : the MICCAI 2009 challenge database (Radau et al., 2009) from the Sunnybrook Health Sciences Center, Toronto, Canada. It is open source and contains three datasets (train, online, validation) of MRI images with 15 cases (from different pathologies). The database also includes manually delineated contours done by experts corresponding to the images. Obviously, the images are in grayscale and we trained our model on the training set. \\

The Figure~\ref{fig:ref} below shows an example of an input image, the inferred region of interest (ROI), corresponding to the manually delineated contours.

\begin{figure}[!htb]
\centering
\subfloat[Input image\label{fig:test1}]
  {\includegraphics[width=.25\linewidth]{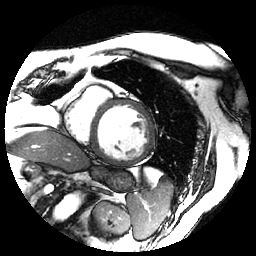}}\hspace{0.5cm}
\subfloat[ROI drawn on image\label{fig:test3}]
{\includegraphics[width=.25\linewidth]{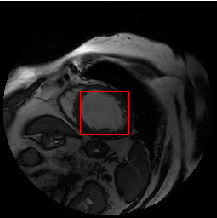}}\hspace{0.5cm}
\subfloat[Contour drawn on image\label{fig:test2}]  
  {\includegraphics[width=.25\linewidth]{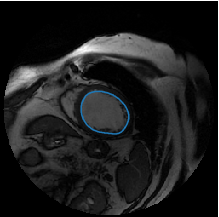}}

\caption{Input image and contour from the the MICCAI 2009 challenge database}
\label{fig:ref}
\end{figure}

\begin{figure}[!htb]
\centering
\subfloat[ROI binary mask \label{fig:test1}]
  {\includegraphics[width=.25\linewidth]{input_mask_roi}}\hspace{0.5cm}
\subfloat[Contour binary mask\label{fig:test2}]
  {\includegraphics[width=.25\linewidth]{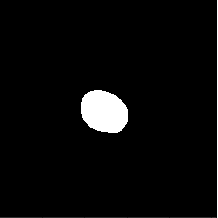}}
\caption{Binary masks from the input image and the corresponding expert contour}
\label{fig:masks}
\end{figure}

From one original MRI image and its corresponding contour (provided as a list of coordinates of every point of the contour), we have to compute binary masks. We use the previous as inputs to our combined approach. An example is shown in Figure~\ref{fig:masks}.

\subsection{Method}
We can divide the approach in three parts (each part uses the output of the previous one):
\begin{enumerate}
\item Left Ventricle Region of Interest extraction using a CNN
\\\textit{Inputs : MR images, Outputs : ROI binary masks}
\item Left Ventricle Inferred shape using a Stacked AutoEncoder
\\\textit{Inputs: ROI, Outputs : Infered shape binary masks}
\item Accurate segmentation using Deformable Models
\\\textit{Inputs: Inferred shape binary masks, Outputs : Contours}
\end{enumerate}
\label{gen_inst}
\subsubsection{Region of Interest}

The first step of our combined approach is to localize the left ventricle on the MR images, i.e draw a bounding box around the LV.
In order to generate this Region of Interest of the input image, we use a Convolutional Neural Network. CNN combine three architectural ideas to ensure some degree of shift and distortion invariance: local receptive fields to extract elementary visual features (edges, corners, etc.), shared weights to perform the same operation on different parts of the image and spatial subsampling to reduce the resolution of the feature map and the sensitivity of the output to shifts and distortions. Since all the weights are learned with back-propagation, convolutional networks can be seen as synthesizing their own feature extractor that enables them to understand complex and interesting shape (such as organs or tissues) to segment the LV.\\

\begin{figure}[H]
\centering
\includegraphics[ width=\linewidth,keepaspectratio]{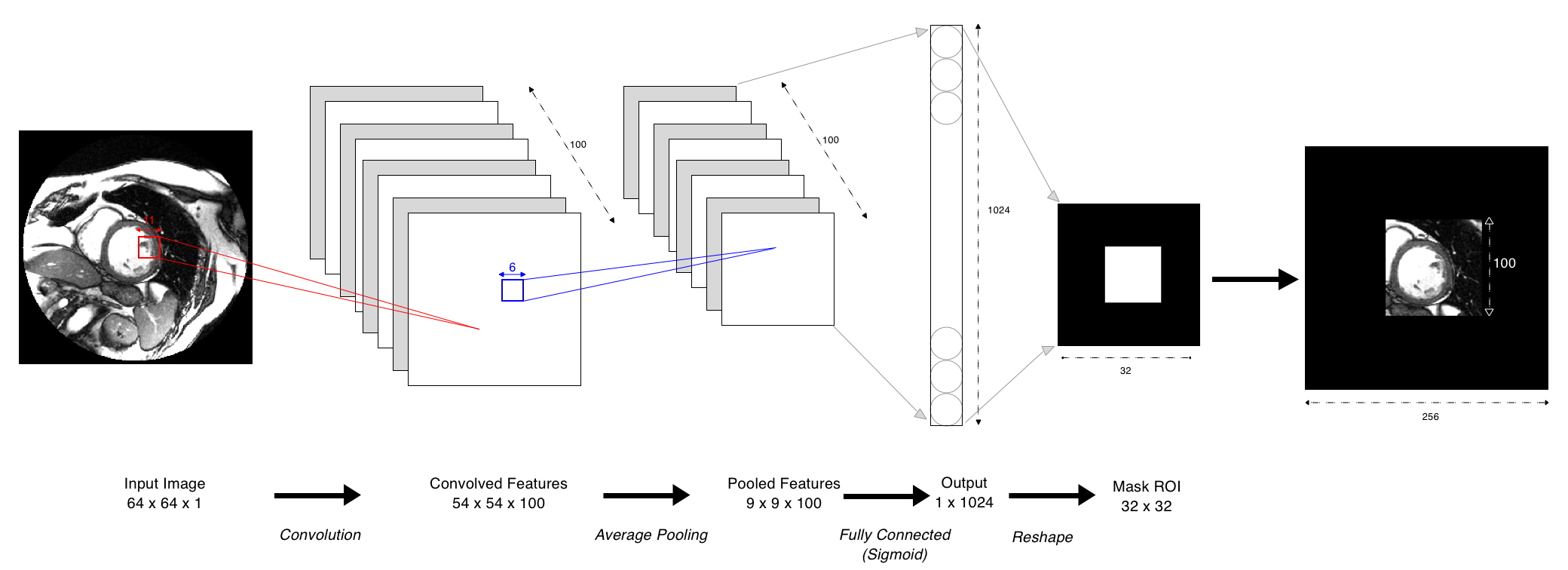}
\caption{Convolutional Neural Network architecture}
\label{fig:cnn}
\end{figure}

The CNN algorithm is represented in the block diagram Figure~\ref{fig:cnn}. As we can see, we only have one convolution layer, followed by a subsampling layer (Average pooling) and a fully connected one computing a logistic regression in order to assign a class to every pixel.
\\

To build the convolution layer, we have 3 degrees of freedom regarding the filters. Indeed, using the notation $W_2$ for the output volume size, $W_1$ for the input volume size (64 in our case as we downsampled the image from $256\times256$ to $64\times64$), 
$F$ for the filter size, $P$ for the padding and $S$ for the stride,the following formula heps us choose our parameters : \begin{align*}
W_2=\frac{W_1-F+2P}{S}+1
\end{align*}
As the LV is often the main part of the input image, we use large kernels ($F_l\in \mathbb{R}^{11\times11}$). The resulting convolved features map has the shape $54\times54\times100$. The pooled features are unrolled from $9\times9\times100$ to $8100\times1$ before the fully connected layer (logistic regression) and then re-rolled to get the ROI binary mask (from $1024\times1$ to $32\times32$). We use this binary mask to crop a $100\times100$ ROI on the $256\times256 $ original MRI image.\\

The convolutional neural network supervised training consists in getting the optimal values for the filters $F_l$, the bias and the weights for the logistic regression. Generally, to train CNN, we need a lot of data but because delineated contour is time consuming and can be done only by experts, we have too few data. To solve this issue, we need a pre-training step for $F_l$ and thus we use a sparse auto-encoder to initialize the filters [2].\\

\newpage
\subsubsection{Inferred shape}

Now that we have retrieved the region of interest in the MRI images, the next step is to find the inferred shape of the LV in it. In order to do so, we used Stacked Auto-encoder. "A stacked auto-encoder is a neural network consisting of multiple layers of sparse auto-encoders in which the outputs of each layer is wired to the inputs of the successive layer" ([3]). The advantage of using such an architecture is to learn new representations of the input image being the ROI and thus locate first-order features (edges, contours), second-order features (patterns, correlations in the appearance of first-order features) and so on.\\

Concretely, we want to locate the exact contour of the LV in the bounding box obtained using CNN. In order to do so, we train a stacked AutoEncoder following two steps : an unsupervised learning step for the training of two sparse auto-encoders that are stacked (the second AutoEncoder uses the output of the first AutoEncoder hidden layer); and a supervised learning step for the training of a fully connected layer (using the output of the second AutoEncoder hidden layer). The architecture is represented in the block diagram Figure~\ref{fig:sae}. 

\begin{figure}[!htb]
\centering
\includegraphics[ width=\linewidth,height = 10.5 cm, keepaspectratio]{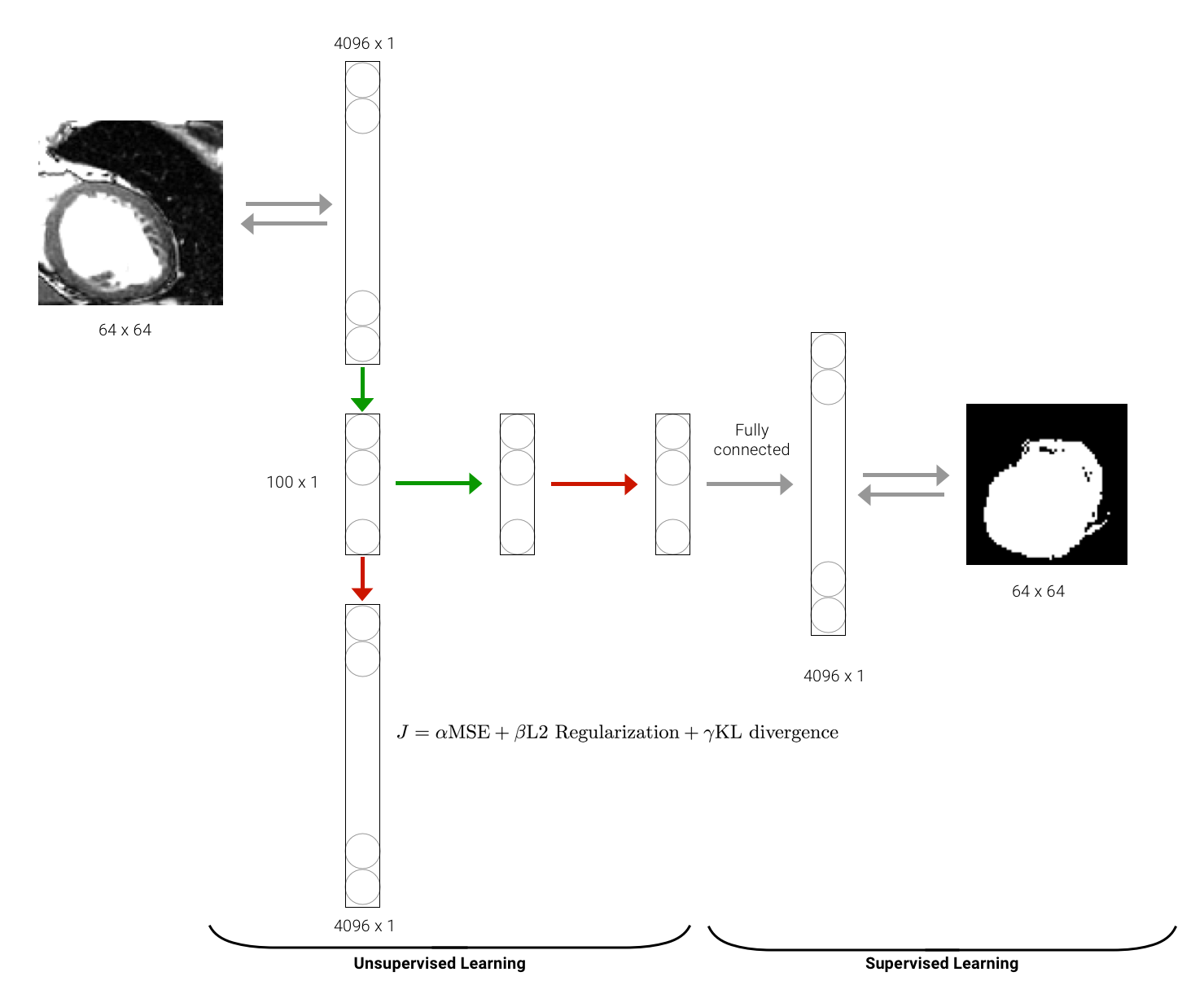}
\caption{Stacked Autoencoder architecture}
\label{fig:sae}
\end{figure}

Thus, the stacked auto-encoder we are using has one input layer (the ROI) of shape $100\times100$ sub-sampled to $64\times64$ and unrolled to $4096\times1$; two hidden layers of size $100\times1$ each; and one output layer of shape $4096\times1$ which is re-rolled in order to obtain the binary mask of the contour of shape $64\times64$.\\

For the unsupervised learning part, the objective is to minimize the loss of the model which is, in our case, a customized loss that combines Mean Square Error (MSE), $L_2$-regularization and Kullback Leibler (KL) divergence. The use of such a loss allows avoiding overfitting ($L_2$-regularization in order to decrease the magnitude of the weights) as well as learning higher representations of the Region of interest (sparsity constraints using KL divergence). \\
Concerning the supervised learning task, we train a fully connected layer using the last representation of the Region of Interest (output of second AutoEncoder) and predict the inferred shape using manual delineation done by experts (ground truth).

\subsubsection{Deformable models}
The ultimate objective is to get the accurate segmentation using dynamic contours, considering the issue of usual shrinkage, leakage \& 3D misalignment. Taken that we have information on both the inferred shape and intensity, we can use Deformable models and then quadratic polynomials for the 3D reconstruction.
\begin{enumerate}[label=(\roman*)]
\item Deformable models are curves which deform under the influence of internal and external forces to delineate exact object boundary, the LV in our case [4]. The internal forces measure the smoothness of the contour while the external forces quantify the distance of the inferred shape towards the LV. In addition, we incorporate the prior shape energy in order to prevent from shrinkage/leakage. Deformable models are classified into two general approaches, the parametric and the geometric models. We will concentrate on the second.\\

\textbf{Geometric or implicit models} use the level-set based shape representation. We denote by $\phi$ the level set function such that :
\begin{align*}
    \phi(X) =
  \begin{cases}
  0, \qquad X \in C \\
  -\min_{X_C}\|X-X_C\|, \qquad X \in R_C\\
  +\min_{X_C}\|X-X_C\|, \qquad X \not\in R_C
    \end{cases}
\end{align*}
where X = (x,y) are the coordinates of image pixel, C is the contour and $R_C$ the region enclosed by C. Thus, the energy function - which is a combination of length-based, region-based and prior shape energy - is such that : 
\begin{align*}
\boxed{E(\phi) = \alpha_1 E_{len}(\phi) + \alpha_2 E_{reg}(\phi) + \alpha_3 E_{shape}(\phi)}
\end{align*}

The unique contour $C^*$ is then obtained by minimizing the energy function using gradient descent initialized with the inferred shape of the previous step: :  
\begin{align*}
\phi^* = \arg \min_{\phi} E(\phi)
\end{align*}
\vspace{0.5cm}
\item We often face 3D misalignment issues due to respiratory and patient motions during MRI scans. \textbf{Quadratic polynomials} are one way of reconstructing continuous surfaces in 3D (contour alignment between slices for 3D reconstruction). We assume that the misalignment values can be modeled as a gaussian noise such that :
\begin{align*}
    \widetilde{x_i} = x_i + w_i \\
    \widetilde{y_i} = y_i + v_i
\end{align*}
where ($x_i$,$y_i$) are the coordinates for the actual center of the ith contour $w_i \sim \mathbb{N} (0, \sigma_{w} ^2)$ and $v_i \sim \mathbb{N}(0,\sigma_v^2)$. Hence, using quadratic polynomials on the 2D cuts from the MRI, we only need to find 6 parameters (three for $x_i$ and three for $y_i$) when minimizing the MSE. \\
Finally, \textbf{affine registration} with linear interpolation provide an aligned stack of contours and hence it results in the accurate contour (LV segmentation).
\end{enumerate}


\newpage
\section{Implementation details}
\subsection{Implementation and metrics}
In order to evaluate our model, we are using the online dataset of the MICCAI database which contains 534 MRI cuts and its corresponding annotated contours.\\

To fully understand the paper, we implemented multiple parts of the explained approach. To this end, we used Keras (deep learning library) with TensorFlow backend and OpenCV in our Python Jupyter notebooks. First, we implemented the convolutional neural network but without the initialization. Indeed, to prevent the lack of data, the approach uses a sparse auto-encoder to initialize the filters so we may face overfitting and decreased performance in comparison with the paper. Also, We did not use any metric to evaluate the performance of the CNN, we stopped the training when the MSE loss converhad ged. Then, we have fully implemented the stacked auto-encoder (the three parts) with its customized loss to predict the inferred shape from the ROI (output of the CNN). And finally, we used the snake algorithm to generate active contours.\\

To measure the performance of the stacked auto-encoder (and thus the combined model CNN and stacked AE) and the final segmentation, we implemented the dice metric and the conformity coefficient, those are the metrics used in the paper to evaluate and finetune the model [5] :
\begin{itemize}
    \item The Dice Metric  : a measure of contour overlap between automatically and manually segmentation
    \begin{align*}
       \boxed{ DM = \frac{2 |A_{a} \bigcap A_{m}|}{|A_{a}|+|A_{m}|}}
    \end{align*}  with $A_{a}$ the area automatically segmented, $A_{a}$ the area manually segmented and $A_{am}$ the intersection
    \item The Conformity Coefficient : the ratio of mis-segmented pixels to the number of correctly segmented ones \begin{align*}
        \boxed{CC = \frac{3DM-2}{DM}}
    \end{align*}
    \item The Average Perpendicular Distance : distance from the predicted contour to the manually drawn, average over all contour points and that we did not have time to implement
    \begin{align*}
        \boxed{APD = \frac{1}{N} \sum_i \| x_i-p(x_i) \|_2}
    \end{align*} with $p(x) = \text{projection of }x\text{ on } C_1$ and 
        $C_1 = \{x_i,   i \in [1, ..., N]\}$

\end{itemize}

Thus we evaluated the model that we have partially implemented and achieved the performance described below in Table~\ref{tab1}. As we could have expected, we get lower performance - considering that we did not implement the complete method. We indeed ignored the initialization of the kernels using a sparse Auto-Encoder for the CNN. What's more, we used the snake algorithms for the active contours and not the level set approach as described in the paper.

\vspace{0.3 cm}
\begin{table}[!htb]
    \centering
\begin{tabular}{||c c c c||} 
 \hline
  & Train set & Validation set & Paper performance \\ [0.5ex] 
 \hline
 Dice Metric & 54.8 \% & 51.6 \% & 89\%\\
 \hline
 Conformity metric & -0.65 & -0.88 & 0.76\\
 \hline
\end{tabular} \vspace{0.2 cm}
    \caption{Dice and conformity metrics for the train and validation sets, in comparison with the paper}
    \label{tab1}
\end{table}

\newpage
\subsection{Results}
\subsubsection{Challenging the CNN part}
We have implemented a model that can segment the LV (inferred and smoothed shape). We decided to challenge the paper on the depth, the width of the filters, the pooling and the activation based on the following intuitions:
\begin{itemize}
    \item The CNN as it is described in the method is not "deep" because we use only one layer of convolution. Overwhelming empirical evidence as well as intuition indicates that having depth in the neural network could be indeed important (Szegedy et al. [6]).
    \item As well as going deeper, we could also suppose that using larger filters would improve the performance.
    \item the Rectified Linear Unit (ReLU) has become very popular in the last few years. It computes the function $f(x)=max(0,x)$. It was found to greatly accelerate (Krizhevsky et al. [7]) the convergence of stochastic gradient descent compared to the sigmoid functions.
    \item A pooling operator operates on individual feature channels, coalescing nearby feature values into one  and reducing the sensitivity of the output to shifts and distortions. In this paper, we are using average pooling although the most common choice is to use max pooling. [8]
\end{itemize}

\begin{figure}[!htb]
\centering
\subfloat[Ground Truth\label{fig:test2}]  
  {\includegraphics[width=.25\linewidth]{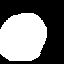}}\hspace{0.5cm}
\subfloat[Original\label{fig:classic}]
  {\includegraphics[width=.25\linewidth]{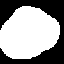}}\hspace{0.5cm}
\subfloat[Deeper\label{fig:deeper}]
{\includegraphics[width=.25\linewidth]{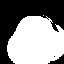}}\hspace{0.5cm}
\subfloat[Larger\label{fig:test2}]  
  {\includegraphics[width=.25\linewidth]{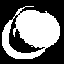}}\hspace{0.5cm}
\subfloat[MaxPooling\label{fig:test2}]  
  {\includegraphics[width=.25\linewidth]{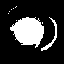}}\hspace{0.5cm}
\subfloat[ReLu\label{fig:test2}]  
  {\includegraphics[width=.25\linewidth]{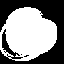}}

\caption{Comparison of the predictions for one MRI cuts of the LV (from the validation dataset) with modified CNN architectures}
\label{fig:compar}
\end{figure}

The Dice Metric (and the conformity coefficient) is our measure of performance on the online dataset that we are using as a validation dataset. Figure~\ref{fig:compar} shows the inferred shape (contour mask) when using a deeper model (two convolutional layers instead of one), a larger model (200 filters instead of 100), a model with ReLu activations and a last one with max pooling (instead of average pooling).

Contrary to our prior intuitions, only a deeper neural network improves slightly the performance of the CNN with respect to the original model. 
\\
\\
\\
\subsubsection{Challenging the Stacked Auto-Encoder part}
Previously, we have modified the parameters of the CNN architecture explained in the paper, without changing the ones from the stacked Auto-Encoder, and we noted the differences on the contour masks. Now, we will use the CNN from the method and only modify the parameters of the stacked auto-encoder in order to assess the performance of generation of the inferred shape.\\

First, we have tried to edit the initialization of the weights of the neural network. Initially, we were initializing the weights tensor to zero. Hence, we have tried with weights initialized with a normal distribution (mean null and a small std.) and with a uniform distribution ($\pm0.05$). The original zero initialization is indeed the best choice (test on the validation set).\\

Customizing the loss of our neural network can totally change the performance of our approach in both ways [9]. Let's recall that the method uses a customized loss for the two stacked autoencoder (weighting between Mean Square Error (MSE), $L_2$ regularisation and Kullback-Leibler (KL) divergence) in the unsupervised learning step and a regular MSE for the last fully connected layer in the supervised learning step. We test another alternative where all the losses are MSE and plot the contours from the manually segmented LV, the prediction with and without active contour models in Figure~\ref{fig:sae}. As we can see, the contour is smoother and seems to fit the ground truth while having a meaningful shape (not shrinking or leaking or irregularities).  

\begin{figure}[!htb]
\centering
\subfloat[Learning curves\label{fig:test1}]
  {\includegraphics[width=.41\linewidth]{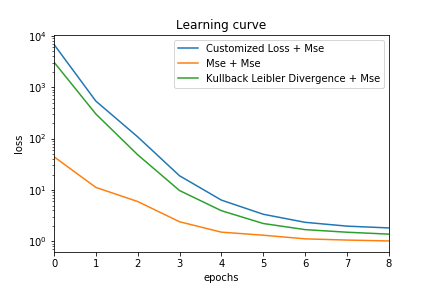}}\hspace{0.5cm}
\subfloat[Predictions\label{fig:test2}]
  {\includegraphics[width=.25\linewidth]{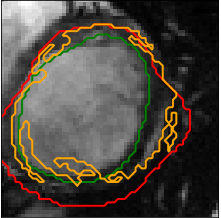}}
\caption{Learning curves and predictions (\textit{customized and MSE losses \textit{(red)}, only MSE losses \textit{(orange)} and ground truth \textit{(green)}} of the segmentation of the LV}
\label{fig:sae}
\end{figure}

\begin{figure}[!htb]
\centering
\includegraphics[width=.25\linewidth]{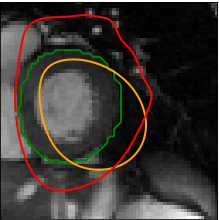}
\caption{A MRI image including : prediction with active contour models on the model with the original loss \textit{(red)}, on the model with only MSE losses \textit{(orange)} and ground truth \textit{(green)} of the segmentation of the LV}
\label{fig:ac}
\end{figure}

Even if we did not implement the geometric algorithm (level-set method) of the Deformable Models, we used the snakes approach with active contour models [10]. We thus smoothed the contours for both the prediction using customized and MSE losses and the one using only MSE losses. The prediction we finally achieve can be very much encouraging with the modified loss of the stacked Auto-Encoder as it is illustrated in Figure~\ref{fig:ac}.
\vspace{0.5cm}

\newpage
\section{Conclusions}
To conclude, let us recall that the goal of the project was to deeply understand the paper and implement parts of the method. The paper presented a unique approach combining Deep Learning and Deformable Models to segment the LV (region-based segmentation), palliating the lack of resources for manual delineation. The method can be parceled in three blocs, each one using the output of the previous as input for the training : the extraction of the Region of Interest using CNN, the inferred shape of the LV using stacked Auto-Encoder and finally the accurate segmentation with Deformable Models and Affine Registration. \\

We mainly implemented the first two blocs and the snakes algorithm for shape modeling. The results we achieve are lower than the ones from the paper because we ommited some parts of the paper, such as the initialization, the dynamic contours using the level set approach, etc. We thus tried to challenge the architecture and the parameters for both the CNN and the Stacked Auto-Encoder and we were quite impressed with the relative increase of performance (deeper CNN + All MSE).\\

Lastly, we were pleased by the paper, its unique method and the approach they adopted to explain each and every choice they made. We found ourselves wanting to get even deeper at each step and ended up implementing the majority of it while challenging the architecture and assumptions.


\section*{References}
[1] M. R. Avendi, A. Kheradvar, and H. Jafarkhani. A Combined Deep-Learning and Deformable-
Model Approach to Fully Automatic Segmentation of the Left Ventricle in Cardiac MRI. ArXiv e-prints, December 2015.

[2]  Andrew Ng. Sparse autoencoder. CS294A Lecture notes.

[3] Chuan Yu Foo Yifan Mai Caroline Suen Andrew Ng, Jiquan Ngiam.  Stacked autoencoder.
Unsupervised Feature Learning and Deep Learning
\url{http://ufldl.stanford.edu/wiki/index.php/Stacked_Autoencoders}.

[4] Xiaolei Huang and Gavriil Tsechpenakis. Deformable model-based medical image segmentation.
In Multi Modality State-of-the-Art Medical Image Segmentation and Registration Methodologies 
pages 33-67. Springer, 2011.

[5] P. Radau, Y. Lu, K. Connelly, G. Paul, A. Dick, and G. Wright.  Evaluation framework for
algorithms segmenting short axis cardiac mri. Jul 2009.

[6] C. Szegedy, W. Liu, Y. Jia, P. Sermanet, S. Reed, D. Anguelov, D. Erhan, V. Vanhoucke, and
A. Rabinovich. Going Deeper with Convolutions. ArXiv e-prints, September 2014.

[7] Alex Krizhevsky, Ilya Sutskever, and Geoffrey E Hinton.  Imagenet classification with deep
convolutional neural networks. In F. Pereira, C. J. C. Burges, L. Bottou, and K. Q. Weinberger,
editors,Advances in Neural Information Processing Systems 25, pages 1097-1105. Curran Associates, Inc., 2012.

[8] Y-Lan Boureau, Nicolas Le Roux, Francis Bach, Jean Ponce, and Yann LeCun. Ask the locals:
multi-way local pooling for image recognition. In Proc. International Conference on ComputerVision (ICCV 11). IEEE, 2011.

[9] K. Janocha and W. M. Czarnecki. On Loss Functions for Deep Neural Networks in Classification.
ArXiv e-prints, February 2017.

[10] Michael Kass, Andrew Witkin, and Demetri Terzopoulos.  Snakes:  Active contour models.
International Journal of Computer Vision, Jan 1988.

\end{document}